\documentclass[sigconf]{acmart}
\AtBeginDocument{%
  \providecommand\BibTeX{{%
    \normalfont B\kern-0.5em{\scshape i\kern-0.25em b}\kern-0.8em\TeX}}}

\setcopyright{acmlicensed}
\copyrightyear{2018}
\acmYear{2018}
\acmDOI{XXXXXXX.XXXXXXX}

\acmConference[Conference acronym 'XX]{Make sure to enter the correct
  conference title from your rights confirmation emai}{June 03--05,
  2018}{Woodstock, NY}
\acmISBN{978-1-4503-XXXX-X/18/06}

\usepackage{booktabs}
\usepackage{diagbox}
\usepackage{multirow}
\usepackage{algorithm} %format of the algorithm
\usepackage{algorithmic} %format of the algorithm
\usepackage{multirow} %multirow for format of table
\usepackage{amsmath}
\usepackage{xcolor}
\theoremstyle{plain}
\newtheorem{theorem}{\noindent \textbf{Theorem}}[]

\newcommand{\sys}{\textsc{TTReFT}}

\begin{document}

\title{Beyond Parameter Finetuning: Test-Time Representation Refinement for Node Classification}

\author{Jiaxin Zhang}
\orcid{0000-0003-1309-5865}
\affiliation{%
  \institution{National University of Defense Technology}
  \streetaddress{Deya road No.109}
  \city{ChangSha}
  \state{Hunan}
  \country{China}
  \postcode{410073}
}
\email{zhangjiaxin18@nudt.edu.cn}

\author{Yiqi Wang}
\orcid{0000-0001-9594-1919}
\authornote{Corresponding Author}
\affiliation{%
  \institution{National University of Defense Technology}
  \streetaddress{Deya road No.109}
  \city{ChangSha}
  \state{Hunan}
  \country{China}
  \postcode{410073}
}
\email{yiq@nudt.edu.cn}

\author{Siwei Wang}
\affiliation{%
  \institution{Intelligent Game and Decision Lab}
  \city{Beijing}
  \country{China}
}
\email{wangsiwei13@nudt.edu.cn}

\author{Xihong Yang}
\author{Yu Shi}
\orcid{0009-0009-5990-0381}
\author{Songlei Jian}
\affiliation{%
  \institution{National University of Defense Technology}
  \streetaddress{Deya road No.109}
  \city{Changsha}
  \country{China}}

\author{Xinwang Liu}
\authornotemark[1]
\affiliation{%
  \institution{National University of Defense Technology}
  \streetaddress{Deya road No.109}
  \city{ChangSha}
  \state{Hunan}
  \country{China}
  \postcode{410073}}
\email{xinwangliu@nudt.edu.cn}

\author{En Zhu}
\authornotemark[1]
\affiliation{%
  \institution{National University of Defense Technology}
  \streetaddress{Deya road No.109}
  \city{ChangSha}
  \state{Hunan}
  \country{China}
  \postcode{410073}}
\email{enzhu@nudt.edu.cn}
\renewcommand{\shortauthors}{Jiaxin Zhang and Yiqi Wang, et al.}




\received{20 February 2007}
\received[revised]{12 March 2009}
\received[accepted]{5 June 2009}


\begin{abstract}
Graph Neural Networks frequently exhibit significant performance degradation in the out-of-distribution test scenario. 
While test-time training (TTT) offers a promising solution, existing Parameter Finetuning (PaFT) paradigm suffer from catastrophic forgetting, hindering their real-world applicability. We propose \sys, a novel \textbf{T}est-\textbf{T}ime \textbf{Re}presentation \textbf{F}ine\textbf{T}uning framework that transitions the adaptation target from model parameters to latent representations. Specifically, TTReFT achieves this through three key innovations: (1) uncertainty-guided node selection for specific interventions, (2) low-rank representation interventions that preserve pre-trained knowledge, and (3) an intervention-aware masked autoencoder that dynamically adjust masking strategy to accommodate the node selection scheme. 
Theoretically, we establish guarantees for \sys~in OOD settings. Empirically, extensive experiments across five benchmark datasets demonstrate that \sys~achieves consistent and superior performance. Our work establishes representation finetuning as a new paradigm for graph TTT, offering both theoretical grounding and immediate practical utility for real-world deployment.
\end{abstract}
\maketitle
\section{Introduction}
% Graph Neural Networks (GNNs) extend Deep Neural Networks to graph data through message passing, achieving success in applications like social network analysis~\cite{hamilton2017inductive} and recommendation systems~\cite{yang2025dual, rossi2021knowledge}. However, their empirical success is heavily based on the critical but often violated IID assumption (independent and identically distributed) between training and test data~\cite{hendrycks2019benchmarking, mancini2020towards}. In practical scenarios, test data frequently exhibit distributional shifts in node features (e.g., attribute changes) or graph structure (e.g., homophily variations). These out-of-distribution (OOD) scenarios lead to significant performance degradation that impedes practical applications. 

Graph Neural Networks (GNNs)~\cite{xu2018powerful} are the successful extension of Deep Neural Networks (DNNs)~\cite{lecun2015deep} on graph data, and have achieved remarkable revolutions in graph-related applications including social network analysis~\cite{hamilton2017inductive} and recommendation systems~\cite{yang2025dual, rossi2021knowledge}. Like DNNs, most GNNs are built on the basis of the IID assumption (independent and identically distributed) between training and test data~\cite{hendrycks2019benchmarking, mancini2020towards}. However, the IID assumption is often violated in practical scenarios. Specifically, graphs can exhibit distributional shifts in either node features (attribute changes) or graph structure (homophily variations), which can lead to significant performance degradation and thus hinder practical applications.
\begin{figure*}[t]
    \centering
    \includegraphics[width=0.9\linewidth]{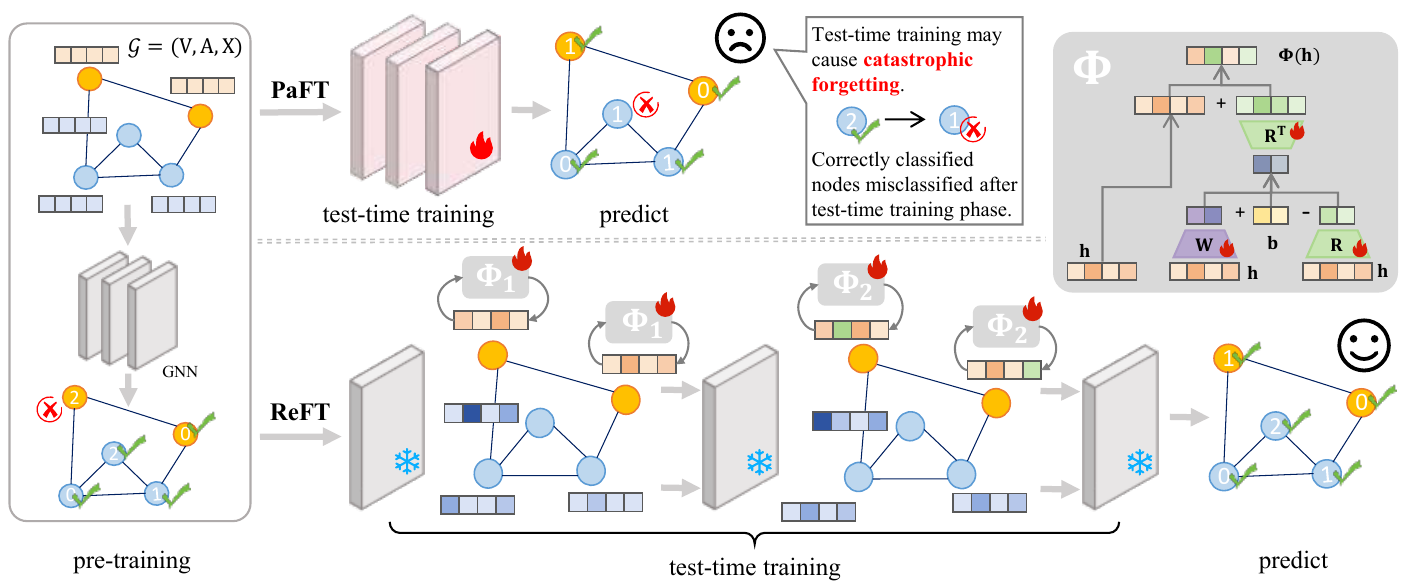}
    \caption{A comparison of the working mechanisms of PaFT and ReFT on graphs. (1) Parameter Finetuning (PaFT): The model parameters are updated during test-time training, which can lead to catastrophic forgetting, as illustrated by correctly classified training samples (blue nodes) being misclassified (turning red) after adaptation. (2) Representation Finetuning (ReFT): The pre-trained model parameters remain frozen. Adaptation is achieved by applying targeted, learnable interventions to a sparse subset of node representations (highlighted in yellow), successfully correcting misclassifications without forgetting pre-trained knowledge.}
    \label{fig:PaFT}
\end{figure*}

Numerous studies have been proposed to address the aforementioned out-of-distribution (OOD) challenges~\cite{li2022out}, with prominent approaches including data augmentation to diversify training distributions~\cite{wu2022knowledge}, adversarial training for perturbation invariance ~\cite{xu2019topology, feng2019graph} and invariance-based causal feature learning~\cite{song2022learning}. These methods typically require multi-domain training data or rely on strong assumptions to synthesize diverse training samples that simulate unknown test features. Some even necessitate partial test data and labels for model adaptation~\cite{feng2020graph}. Such requirements and assumptions are often infeasible in practice, thereby hindering their deployment. Moreover, these methods often fail to adequately explore and leverage the distributional characteristics of test data. To mitigate these issues, Test-Time Training (TTT)~\cite{sun2020test} has emerged as a novel paradigm in recent years. TTT typically employs unsupervised learning to dynamically adapt a subset of model parameters, enabling the model to better align with the distribution of test data. Multiple studies have demonstrated its effectiveness and potential in addressing OOD challenges~\cite{zhang2024test,wang2022test}. However, prevailing TTT approaches primarily rely on Parameter Finetuning (PaFT), a paradigm that struggles to ensure stable performance but may even induce severe problems such as catastrophic forgetting~\cite{liu2021ttt++}.

Recent advances in representation learning suggest a novel promising solution for test-time training~\cite{wu2024reft, chen2023sudden}. Specifically, the success of Representation Finetuning (ReFT) in Natural Language Processing (NLP)~\cite{wu2024reft} demonstrates that targeted representation interventions can match full finetuning performance while modifying less than 1\% of representations. Furthermore, theoretical work has established that neural networks inherently organize hierarchical semantic information within their representation spaces~\cite{geiger2021causal}, and suggests that conventional parameter updates may disrupt these patterns.
Motivated by these insights, we introduce the ReFT paradigm to the graph domain as a superior alternative to PaFT. As schematically compared in Figure~\ref{fig:PaFT}, unlike PaFT which updates global model weights, the general ReFT paradigm keeps all pre-trained parameters frozen to prevent catastrophic forgetting. Instead, it learns targeted, low-rank interventions on hidden representations for efficient node-wise adaptation.
However, seamlessly transplanting ReFT to graph data is non-trivial and faces two unique challenges: (1) Unlike NLP where interventions can target specific tokens, graph nodes are interconnected, raising the question of \textit{which nodes should be selected for intervention?}; (2) Given the absence of test labels, \textit{how can we guide this adaptation effectively?}

% Building upon this general paradigm, we propose \sys, a novel \textbf{T}est-\textbf{T}ime \textbf{Re}presentation \textbf{F}ine\textbf{T}uning framework specifically designed for graph data.
% However, applying ReFT to graphs presents two unique challenges: \textit{which nodes should be selected for intervention, and how to guide this adaptation effectively without labels?}

% Motivated by these insights, we propose \sys, a novel \textbf{T}est-\textbf{T}ime \textbf{Re}presentation \textbf{F}ine\textbf{T}uning framework. Instead of partially adjusting model parameters, \sys~keeps all pre-trained parameters unchanged to prevent catastrophic forgetting and preserves learned knowledge. Meanwhile, it learns targeted, low-rank interventions on hidden representations for efficient node-wise adaptation. As illustrated in Figure~\ref{fig:PaFT}, this approach enables plug-and-play test-time training by adjusting activations on the fly, offering a computationally efficient and easy-to-integrate solution for adapting models to new data.
% \jt{readers may be confused by Figure 1 and figure 2. seems both figures. As I mentioned later, the ReFT part is a general frameowork while  TTREFT  is an imeplentation?? if this is the case, we may say " a general TTT frmework under ReFT is in figure 1 with a comparions PaFT... then based on the general frmaeowkr, we propose TTREFT.., ....... Pleas also see my comment in Section 2.3 } 

To address these challenges, we propose \sys, a novel \textbf{T}est-\textbf{T}ime \textbf{Re}presentation \textbf{F}ine\textbf{T}uning framework that instantiates ReFT for graph data as illustrated in Figure~\ref{fig:model}.
To tackle the first challenge of node selection, we introduce an uncertainty-guided node selection mechanism. Instead of random selection, our approach prioritizes nodes with high predictive entropy, strategically focusing adaptation resources on uncertain samples most affected by distribution shifts. 
To solve the second challenge of guidance, we design a self-supervised reconstruction task tailored for interventions.
Although contrastive learning is popular~\cite{hassani2020contrastive, qiu2020gcc}, its reliance on artificial augmentations~\cite{zhang2021canonical} may not reflect real-world test-time shifts. Instead, we propose an intervention-aware masked autoencoder (IAMAE). Specifically, we align reconstruction with our adaptation objective: for each selected node, we reconstruct the features of its local neighborhood using the refined representations. This design not only encourages the refined features to fit the test distribution but also leverages local graph topology as a stability regularizer. Consequently, it generates robust self-supervision signals naturally coupled with the test-time structure, avoiding hand-crafted augmentations.

\sys~unifies these solutions into three coordinated components as shown in Figure~\ref{fig:model}: (1) \textbf{Uncertainty-guided node selection} pinpoints high-entropy nodes as primary intervention targets; (2) \textbf{Low-rank representation intervention} applies parameter-efficient transformations solely to these selected nodes, preserving pre-trained knowledge; and (3) \textbf{Intervention-aware reconstruction} dynamically correlates masking probability with intervention density, creating a closed-loop adaptation system. Through these principled representation-space interventions, our method successfully resolves the inherent tension between adaptation and preservation.
Our key contributions can be summarized as follows: 
\begin{itemize}
\item We propose \sys, the first representation finetuning framework designed for graph, establishing a new paradigm for test-time training that shifts from parameter updating to representation intervention.
\item We develop a novel intervention-aware masked autoencoder (IAMAE), which is the first to explicitly couple the masking probability with local intervention density, enabling targeted self-supervision.
\item We provide theoretical insights that demonstrate the rationality and effectiveness of representation finetuning in graph out-of-distribution (OOD) settings.
\item 
%\jt{does PaFT compromise in-distribution performance?? } 
% Empirically, extensive experiments across five datasets demonstrate that \sys consistently improves OOD generalization while preventing catastrophic forgetting.
Through extensive experiments on five datasets, we empirically validate that \sys~consistently enhances OOD generalization while effectively preventing catastrophic forgetting of pre-trained knowledge.
\end{itemize}

\section{Preliminaries}

This section defines the problem scope and contrasts two adaptation paradigms: the conventional Parameter Finetuning (PaFT) and the Representation Finetuning (ReFT). 
\subsection{Problem Formulation}
% We formalize the node classification task on a graph $\mathcal{G}=(V, A, X)$,where $V = \{v_1,...,v_N\}$ denoted the node set and $N$ is the number of nodes, $A \in {\{0,1\} }^{N \times N}$ is the adjacency matrix, and $X \in \mathbb{R}^{N \times d}$ is the input feature matrix \jt{what is d}. 
We formalize the node classification task on a graph $\mathcal{G}=(V, A, X)$, where $V = \{v_1,...,v_N\}$ denotes the node set with $N$ nodes, $A \in {\{0,1\} }^{N \times N}$ is the adjacency matrix, and $X \in \mathbb{R}^{N \times d}$ is the feature matrix with features dimension $d$.

\noindent \textbf{Graph OOD Problem.}  Given a labeled source graph $\mathcal{G}_s$ and unlabeled target graph $\mathcal{G}_t$ with distribution shift, the goal is to adapt a pre-trained GNN $f_\theta$ to generalize well on $\mathcal{G}_t$. The model parameters $\theta$ are trained on $\mathcal{G}_s$ via supervised loss $L_{sup}$.

\noindent \textbf{Graph Autoencoders (GAEs).}
We employ GAEs for self-supervision. A GAE consists of an encoder $f_E$ mapping inputs to latent representation $\mathbf{H} \in \mathbb{R}^{N \times d_h}$, and a decoder $f_D$ reconstructing the graph structure or features. The reconstruction objective is generally formulated as:
\begin{equation}
    % \mathbf{H} = f_E(A, X),~\mathcal{G}' = f_D(A,\mathbf{H})
    \mathbf{H} = f_E(A, X), \quad \hat{X} = f_D(A, \mathbf{H})
\end{equation}
where $\hat{X}$ denotes the reconstructed features.
\subsection{PaFT for TTT on Graphs}
Standard Test-Time Training (TTT) typically follows a Parameter Finetuning (PaFT) pipeline. Given a pre-trained $K$-layer GNN $f_\theta$ with parameters $\theta=(\theta_1,...,\theta_K)$, PaFT adapts the model by:
\begin{enumerate}
\item Fixing the initial $k$ layers $\theta_{1:k}$ (feature extractor);
\item Updating remaining layers $\theta_{k+1:K}$ by minimizing a self-supervised loss $L_{ssl}$ on the test graph $\mathcal{G}_t$;
\item Inferring with the updated parameters $\theta'={(\theta_{1:k},\theta'_{k+1:K})}$.
\end{enumerate}
However, directly updating $\theta$ often leads to catastrophic forgetting of source knowledge.
\subsection{ReFT Paradigm on Graphs}
\label{sec:reft_paradigm}
% \jt{I am confused about this part. this "Test-Time Representation Finetuning Pipeline" is also proposed by this work, right? it is not from other papers, right??  what is the relation between here and the proposed framework?? Is this a general framewokr and the  TTREFT is an implementation of this framework?? if this is the case, we should put this subsection in 3, first intro a general frameowrk and then TTREFT is an implementaton with detailed designs????  }
We formalize ReFT's general application to graph TTT. Distinct from PaFT, the Representation Finetuning (ReFT) paradigm freezes all model parameters $\theta$ and instead learns to intervene on latent representations. 
Let $\mathbf{h}^{(l)}$ denote the node representations at layer $l$. A general ReFT intervention is defined as a tuple $I=\langle\Phi, \mathcal{P}, \mathcal{L}\rangle$:
\begin{itemize}
    \item $\Phi: \mathbb{R}^d \to \mathbb{R}^d$ is a learnable intervention function parameterized by $\phi$.
    \item $\mathcal{P} \subseteq V$ is the set of target nodes to intervene.
    \item $\mathcal{L}$ is the set of layers where interventions are applied.
\end{itemize}
The intervened representation $\tilde{\mathbf{h}}_v$ is computed as:
\begin{equation}
    \tilde{\mathbf{h}}_v^{(l)} \leftarrow \mathbf{1}(v\in \mathcal{P})\cdot\Phi(\mathbf{h}_v^{(l)}) + \mathbf{1}(v\notin \mathcal{P})\cdot\mathbf{h}_v^{(l)}
\end{equation}
where $\mathbf{h}_v^{(l)}$ is the original representation of node $v$, $\tilde{\mathbf{h}}_v^{(l)}$ is the refined representation, and $\mathbf{1}(\cdot)$ is the indicator function.

During test-time adaptation, ReFT optimizes only $\phi$ using $L_{ssl}$ while $\theta$ remains frozen. This paradigm theoretically circumvents catastrophic forgetting by preserving the original model weights.

\section{Methodology}

\begin{figure*}
    \centering
    \includegraphics[width=\linewidth]{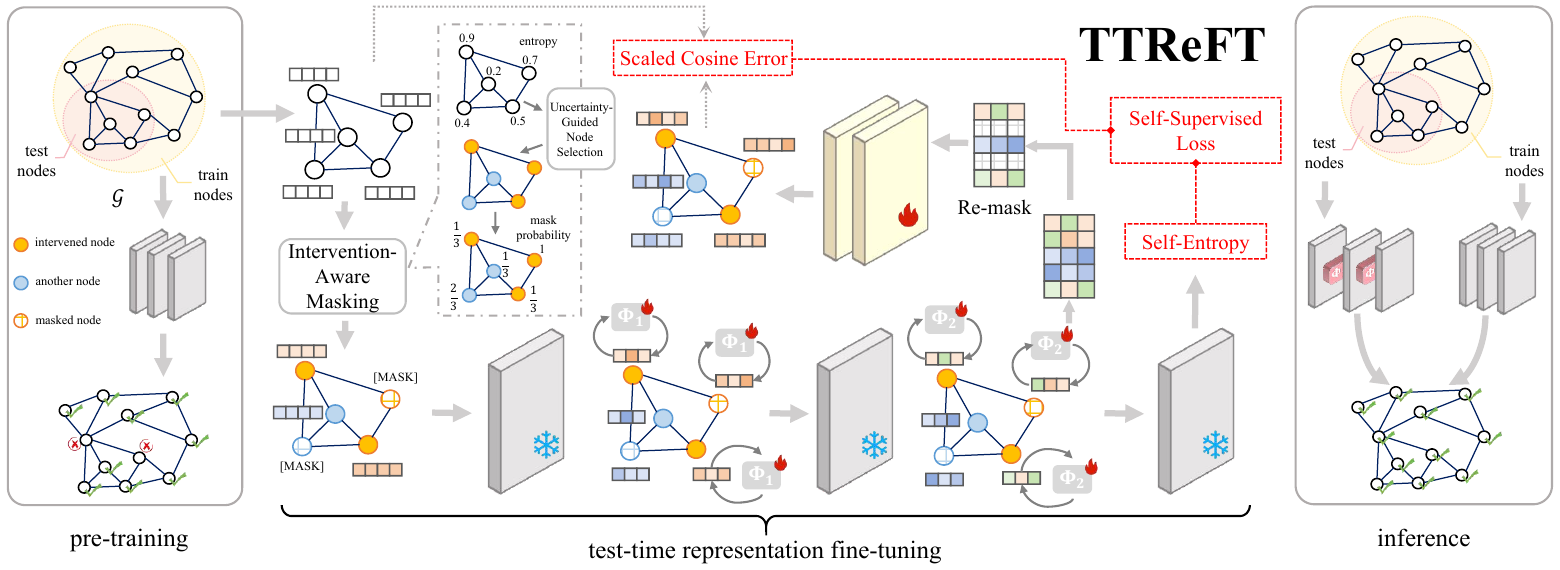}
    \caption{The overall framework of \sys. Our framework operates in three stages: (1) Pre-training: A GNN model is trained on source-domain data. All parameters are frozen after this stage. (2) Test-Time Representation Finetuning: For unlabeled test data, nodes with high predictive uncertainty (highlighted in yellow) are selected. A learnable Low-Rank Representation Intervention (LoReFT) is applied to them. The intervention parameters are optimized using a novel Intervention-Aware Masked Autoencoder (IAMAE) loss, which dynamically masks features based on local intervention density. (3) Inference: The frozen pre-trained model and the learned interventions are combined for final prediction, with interventions only applied to the selected uncertain nodes.}
    \label{fig:model}
\end{figure*}
Based on the general ReFT paradigm formalized in Section~\ref{sec:reft_paradigm}, our proposed \sys framework implements an efficient test-time adaptation strategy specifically tailored for graph data. By shifting the focus from parameter updates to representation interventions, \sys~inherently circumvents catastrophic forgetting while enabling precise local adaptation.
\subsection{Framework Overview}
As illustrated in Figure~\ref{fig:model}, \sys~freezes the pre-trained GNN $f_\theta$ to preserve source knowledge and introduces a lightweight, plug-and-play intervention module. The adaptation process is driven by three coordinated components that instantiate the defined intervention tuple $I=\langle\Phi, \mathcal{P}, \mathcal{L}\rangle$:
(1) \textbf{Uncertainty-Guided Node Selection} determines the target set $\mathcal{P}$ by identifying high-entropy node;
(2) \textbf{Low-Rank Representation Intervention} defines the function $\Phi$ to apply parameter-efficient corrections to the representations of nodes in $\mathcal{P}$;
and (3) \textbf{Intervention-Aware Masked Autoencoders (IAMAE)} provides the self-supervised objective $L_{ssl}$ to jointly optimize the intervention parameters without ground-truth labels.

\subsection{Uncertainty-Guided Node Selection}
\label{Sec:node_selection}
Empirical observations (see Figure~\ref{fig:ablation}(b)) indicate that indiscriminate intervention among all nodes leads to both computational inefficiency and potential degradation of confident predictions. Therefore, we develop a dynamic node selection strategy that prioritizes nodes exhibiting high predictive entropy, which serves as a reliable proxy for distribution shift uncertainty.

For each test node $v_i \in \mathcal{G}_t$, we compute its predictive entropy as $E_i = -\sum_{c=1}^C p(y=c \mid x_i) \log p(y=c \mid x_i)$.
This entropy measure drives a probabilistic intervention decision via a smooth thresholding mechanism:
\begin{equation}
p_i^{\text{intervene}} = \sigma(\alpha_{\mathrm{gate}} (E_i - E_{\text{threshold}})),
\end{equation}
where $\sigma$ is the sigmoid function, $\alpha_{\mathrm{gate}}$ controls the sharpness of the threshold transition and $E_{\text{threshold}}$ is a fixed threshold.. The intervention mask $\mathcal{M} \in \{0, 1\}^{N}$ is then sampled from a Bernoulli distribution parameterized by this probability : $m_i^{\text{intervene}} \sim \text{Bernoulli}(p_i^{\text{intervene}})$. The target node set is thus defined as $\mathcal{P} = \{v_i \in \mathcal{G}_t \mid m_i^{\text{intervene}} = 1\}$. This ensures adaptation resources are concentrated on regions where the model is least confident.

\subsection{Low-Rank Representation Intervention}
To enable efficient adaptation while preserving the essential structure of learned representations, we propose to intervene in a low-rank subspace rather than modifying the full high-dimensional feature space.
For nodes selected in the intervention set $\mathcal{P}$, we employ a Low-Rank Linear Representation Intervention. As shown in Figure~\ref{fig:PaFT}, at target layer $l \in \mathcal{L}$, the intervention function $\Phi^{(l)}$ transforms the original latent representation $h_v^{(l)} \in \mathbb{R}^d$ into a refined representation:
\begin{equation}
\tilde{\mathbf{h}}_v^{(l)} = \mathbf{h}_v^{(l)} + \mathbf{R}^\top (\mathbf{W}\mathbf{h}_v^{(l)} + \mathbf{b} - \mathbf{R}\mathbf{h}_v^{(l)}) 
\end{equation}
where $\mathbf{W} \in \mathbb{R}^{r \times d}$ is the low-rank projection matrix, $\mathbf{b} \in \mathbb{R}^r$ is the bias vector, and $\mathbf{R} \in \mathbb{R}^{r \times d}$ is the subspace basis matrix, with $r \ll d$. The term $\mathbf{W}\mathbf{h} + \mathbf{b} - \mathbf{R}\mathbf{h}$ represents the learned modification within the $r$-dimensional subspace, which is then projected back to the original space via $\mathbf{R}^\top$. To prevent excessive deviation from the pre-trained manifold, we impose an orthogonality constraint on the rows of $\mathbf{R}$ ($\mathbf{R}\mathbf{R}^\top = \mathbf{I}_r$).

Consistent with the general ReFT definition, the intervention is applied only to the selected high-entropy nodes. The final representation used for downstream prediction is given by:
\begin{equation}
    f_{\text{final}}(x_v) = 
    \begin{cases} 
     \Phi^* \circ f_\theta {(x_v)} & \text{if } v \in P, \\
    f_\theta(x_v) & \text{otherwise}.
    \end{cases}
    \label{eq:inference}
\end{equation}
where $\Phi^*$ represents the optimal intervention function learned during test time.

This design offers three key advantages: (1) Parameter Efficiency: By optimizing only $\{\mathbf{W}, \mathbf{b}, \mathbf{R}\}$ where $r \ll d$, the number of tunable parameters is reduced by orders of magnitude compared to full parameter fine-tuning. (2) Structural Preservation: The residual nature of the intervention ($\tilde{\mathbf{h}} = \mathbf{h} + \Delta \mathbf{h}$) ensures that the original semantic information is largely preserved, mitigating catastrophic forgetting. (3) Targeted Adaptation: By restricting updates to the subspace defined by $\mathbf{R}$, we force the adaptation to focus on the most salient directions of distribution shift.

\subsection{Intervention-Aware Masked Autoencoder}
To guide the optimization of the intervention parameters $\phi=\{\mathbf{W}, \mathbf{b}, \mathbf{R}\}$ without labeled target data, we develop an Intervention-Aware Masked Autoencoder (IAMAE). This approach fundamentally extends standard masked autoencoding by coupling the reconstruction task with the intervention density, recognizing that intervened nodes and their neighborhoods contain the most critical information regarding distribution shifts.

\subsubsection{Intervention-aware Masking}
Unlike standard Masked Autoencoders (MAE) that employ uniform random masking, we propose a intervention-aware adaptive masking strategy. Our intuition is that nodes surrounded by a high density of intervened neighbors are likely located in complex decision boundaries or distribution-shifted regions. These nodes should be masked with higher probability to force the model to learn more robust representations from their context.

Specifically, given the intervention mask $\mathcal{M} \in \{0, 1\}^{|\mathcal{V}|}$, for node $v_i$, we calculate its neighbor intervention count $C_i = \sum_{v_j \in \mathcal{N}(v_i)} m_j^{\text{intervene}}$, where $\mathcal{N}{(v_i)}$ denotes the neighborhood of node $v_i$. The masking probability $p_{\text{mask}}^{(i)}$ is dynamically adjusted based on its local intervention density:
% \begin{equation}
%     p_{\text{mask}}^{(i)} = \rho \cdot \left( \beta + (1 - \beta) \cdot \frac{C_i}{\max\limits_{k} C_k} \right)
% \end{equation}
\begin{equation}
        p_{\text{mask}}^{(i)} = \rho \cdot \left( \beta + (1 - \beta) \cdot \frac{C_i}{\max_{k} C_k + \epsilon} \right)
\end{equation}
where $\rho$ is the global masking rate, $\beta$ controls the base masking probability for nodes with no intervened neighbors, and $\epsilon$ is a small constant for numerical stability. This strategy systematically increases the likelihood of masking nodes in regions heavily affected by interventions, forcing the intervention module to generate features that are not only discriminative but also structurally consistent with their neighbors.

\subsubsection{Optimization Objective}
The IAMAE operates on the input graph $\mathcal{G} = (V,A,X)$ by generating masked features $\tilde{X}$ according to $p_{\text{mask}}^{(i)}$. Specifically, we sample a subset of nodes $\tilde{V}$ to mask and replace their features with a learnable $[MASK]$ vector. 
The encoder $f_E$ corresponds to the pre-trained model $f_\theta$ augmented with the intervention module $\Phi$. It takes $\tilde{X}$ and adjacency matrix $A$ as input to produce latent representations $\mathbf{H}$. Crucially, gradients from the reconstruction loss flow through $\mathbf{H}$ to update the intervention parameters $\phi$, while $\theta$ remains frozen.

During decoding, we apply re-masking to the representations before reconstructing the original features through a lightweight decoder $f_D$. This two-stage masking process ensures that the reconstruction task focuses on learning meaningful patterns from the most relevant regions of the graph.
The reconstruction quality is assessed using a Scaled Cosine Error (SCE) loss:
\begin{equation}
L_{\text{IAMAE}} = \frac{1}{|\tilde{V}|} \sum_{v_i \in \tilde{V}} \left(1 - \frac{x_i^\top z_i}{\|x_i\| \|z_i\|}\right)^\gamma,
\end{equation}
where $z_i$ is the reconstructed feature and $\gamma \ge 1$ focuses learning on hard samples.
To further regularize the adaptation, we combine this loss with self-entropy minimization~\cite{shannon1948mathematical}. While IAMAE ensures structural consistency, the entropy loss $\mathcal{L}_e = -\sum_{c=1}^C p_c \log p_c$ promotes confident predictions by minimizing output distribution entropy. The overall self-supervised objective for test-time representation finetuning is:
\begin{equation}
L_{ssl} = L_{IAMAE} + \lambda_{e}L_{e}
\end{equation}
where $\lambda_{e}$ is balancing coefficients.
\subsection{Complexity Analysis}
\sys~significantly reduces computational overhead. For a graph with $N$ nodes and feature dimension $d$, our method applies low-rank interventions (with rank $r$) only to a selected subset of nodes $\mathcal{P}$.
Compared to parameter finetuning which typically scales with $\mathcal{O}(N d^2)$ (updating full weight matrices), our method operates with $\mathcal{O}(|P| r d)$ complexity. Since $|\mathcal{P}| \ll N$ (due to selective intervention) and $r \ll d$ (due to low-rank design), the number of tunable parameters and floating-point operations are reduced by orders of magnitude. This ensures scalability for large-scale graph adaptation. The empirical comparisons to validate this advantage in inference time are included in Table~\ref{Table:efficiency_comparison}.

\section{Theoretical Analysis}

We now demonstrate the theoretical efficacy of representation intervention under distribution shifts. 

\noindent \textbf{Problem Setting.} We consider a node classification task with $C$ classes. We adopt a $K$-layer Simplified Graph Convolution (SGC) model~\cite{wu2019simplifying} for its tractability while retaining graph filtering properties similar to GCNs. The model generates predictions as $Y = \text{softmax}(A^K X W)$, where $W \in \mathbb{R}^{d \times C}$ is the trained weight matrix. We assume the ground-truth labels $Y$ are generated by a 1-layer SGC: $Y = \text{softmax}(A X W_\ast)$ with a fixed underlying weight matrix $W_\ast \in \mathbb{R}^{d \times C}$. 

\noindent \textbf{Algorithms.} At test time, the data undergoes a distribution shift characterized by an orthogonal transformation $Q \in \mathbb{R}^{d \times d}$, yielding test samples $(\tilde{X}, Y) \sim P_t$ where $\tilde{X} = Q X$. To mitigate this, we apply a representation intervention
$\Phi(\tilde{X}) = (1-\alpha)\tilde{X} + \alpha U V \tilde{X}$ (simplification for LoReFT) with fixed $U \in \mathbb{R}^{d \times m}$, $V \in \mathbb{R}^{m \times d}$ ($m \ll d$) and $\alpha \in [0,1]$. Here, $\alpha=0$ corresponds to no adaptation, while $\alpha=1$ represents maximal intervention. Predictions use $\hat{Y}_\alpha = \text{softmax}(A \Phi(\tilde{X}) W)$, with risk defined as:  
\begin{equation}
    \mathcal{R}(\alpha) = \mathbb{E}_{(\tilde{X},y) \sim P_t} \left[ \| \hat{y}_\alpha - y \|_1 \right]
\end{equation}
where $\hat{y}_\alpha$ and $y$ denote row vectors of $\hat{Y}_\alpha$ and $Y$.
 
\begin{theorem}  
\label{theorem}
(Effectiveness of Test-Time Intervention under Orthogonal Shift).  
Let $(X, Y) \sim P$ be training samples with labels generated by $Y = \text{softmax}(A X W)$. Under test distribution $P_t$ induced by $\tilde{X} = Q X$ for orthogonal $Q$, and given an intervention $\Phi(\tilde{X})$ with $U, V$ chosen to counteract $Q$ in the prediction-relevant subspace (see Appendix), there exists $\alpha > 0$ such that:  
\begin{equation}
{Risk}(\alpha) < {Risk}(0).
\end{equation}
\end{theorem}
The complete proof is detailed in Appendix~\ref{Appedix:Proof}.

\noindent \textbf{Remark and Limitations.} Theorem~\ref{theorem} provides the first theoretical justification that representation-space intervention can strictly reduce classification risk under distribution shifts. We acknowledge that our analysis relies on simplifying assumptions to ensure mathematical tractability. However, the insight can be extended to non-linear GNNs and more complex distribution shifts, such as non-orthogonal or topological shifts, as validated by our extensive experiments on the GOOD benchmark.
Extending these theoretical guarantees remains a non-trivial open problem and a promising direction for future research.
% suggests that for distribution shifts, a well-designed representation intervention ($\alpha > 0$) strictly reduces the classification risk compared to no adaptation ($\alpha=0$). This provides the first theoretical justification for test-time representation intervention (as opposed to parameter finetuning) in mitigating distribution shifts for GNNs.
% \begin{table}[t]
% \caption{Node classification accuracy (\%) on test data between \sys and baselines under concept\_degree split. The table is transposed to compare methods across datasets. Bold entries indicate the best performance. "OOM" means out of memory.}
% \label{Table:main}
% \centering
% \setlength{\tabcolsep}{1.5mm}{
% \begin{tabular}{lccccc}
% \toprule
% Method & cora & pubmed & citeseer & wikics & arxiv \\ \midrule
% EERM     & 88.44 & OOM   & 69.30 & 79.89 & OOM   \\
% TAR & 87.74    &  84.74   &  73.79   &  78.84   &  65.79   \\ \midrule
% Tent       & 87.21 & 85.09 & 70.48 & 77.19 & 65.40 \\
% Gtrans     & 85.75 & 79.64 & 69.43 & 77.06 & 63.81 \\
% HomoTTT   & 87.04 & 85.09 & 70.48 & 78.83 & 66.74 \\ \midrule
% \sys     & \textbf{88.77} & \textbf{86.89} & \textbf{75.80} & \textbf{79.96} & \textbf{69.57} \\ \bottomrule
% \end{tabular}}
% \end{table}

\section{Experiment}

This section presents a comprehensive evaluation of the proposed \sys~framework. We first detail the experimental setup, then address the following research questions:
\\ \noindent \textbf{RQ1}. Does \sys~demonstrate superior performance in OOD generalization scenarios?
\\ \noindent \textbf{RQ2}. How well does \sys~maintain original task performance during adaptation?
\\ \noindent \textbf{RQ3}. How does the self-supervised task and node selection strategy affect \sys's performance?  
\\ \noindent \textbf{RQ4}. How sensitive is \sys~to its key parameters?
%\jt{I will not say how to deploy our method which is a very strong claim. I will say: we invesgigate how sensitive of our method is to its key parameters???} 
% What are the optimal configurations for deploying \sys in practical applications?
\begin{table}[t]
\caption{Node classification accuracy (\%) on test data between \sys~and baselines under concept\_degree split. The table is transposed to compare methods across datasets. Bold entries indicate the best performance. "OOM" means out of memory.}
\label{Table:main}
\centering
\fontsize{7.5}{9}\selectfont  % 整体7.5pt
\setlength{\tabcolsep}{1.2mm}{  % 进一步减小列间距
\newcommand{\std}[1]{\raisebox{0.2ex}{\tiny #1}}  % 标准差用\tiny，并微调位置
\begin{tabular}{lccccc}
\toprule
Method & cora & pubmed & citeseer & wikics & arxiv \\ \midrule
EERM     & 88.44\std{ ±0.98} & OOM   & 69.30\std{ ±1.81} & 79.89\std{ ±0.10} & OOM   \\
TAR & 87.74\std{ ±0.23}   &  84.74\std{ ±0.12}   &  73.79\std{ ±0.22}   &  78.84\std{ ±0.14}   &  65.79\std{ ±0.17}   \\ \midrule
Tent       & 87.21\std{ ±0.00} & 85.09\std{ ±0.00} & 70.48\std{ ±0.00} & 78.63\std{ ±0.00} & 65.40\std{ ±0.00} \\
Gtrans     & 85.75\std{ ±0.02} & 79.64\std{ ±0.13} & 69.43\std{ ±0.23} & 75.68\std{ ±0.23} & 63.81\std{ ±0.21} \\
HomoTTT   & 87.04\std{ ±0.00} & 85.09\std{ ±0.00} & 70.48\std{ ±0.00} & 78.89\std{ ±0.00} & 66.74\std{ ±0.00} \\ \midrule
\sys     & \textbf{88.77\std{ ±0.01}} & \textbf{86.89\std{ ±0.00}} & \textbf{75.80\std{ ±0.03}} & \textbf{79.96\std{ ±0.02}} & \textbf{69.57\std{ ±0.00}} \\ \bottomrule
\end{tabular}}
\end{table}

\subsection{Experimental Settings}
\subsubsection{Datasets}
Following the protocol of GOOD~\cite{gui2022good}, we evaluate our method on five established node classification datasets:  cora~\cite{mccallum2000automating}, pubmed~\cite{sen2008collective}, citeseer~\cite{10.1145/276675.276685}, wikics~\cite{mernyei2022wikics} and arxiv~\cite{hu2021open}. To evaluate OOD generalization, we design distinct data splits that explicitly separate covariate shifts (e.g., word distribution changes) from concept shifts (e.g., degree-based distribution changes). Detailed dataset statistics and splitting strategies are provided in Table~\ref{Table: Datasets} in Appendix~\ref{Appendix:Datasets}.
\subsubsection{Evaluation and Implementation}
We adopt the widely used metric accuracy to evaluate the model performance. All experiments were conducted five times using different seeds and the mean performance is reported. The pipeline can be applied to any GNN model, with the most popular GCN~\cite{kipf2017semisupervised} being adopted in this experiment. The results of other GNN model (GAT~\cite{velivckovic2017graph} and SAGE~\cite{hamilton2017inductive}) are detailed in Appendix. Instead of undergoing complex tuning, we fixed the learning rate used in prior studies for all datasets. The code and more implementation details are available in supplementary material.
\subsubsection{Baselines}
We compare \sys~against state-of-the-art approaches categorized into two groups: Graph-specific Domain Generalization (DG) and Test-Time Training. The DG group includes (1) EERM~\cite{wu2022handling}, which employs environment-based regularization, and (2) TAR~\cite{zhengtopology}, a recent SOTA method utilizing topology-aware dynamic reweighting strategies. The TTT group comprises (3) Tent~\cite{wang2020tent}, a general entropy minimization approach adapted from vision; (4) GTrans~\cite{jin2022empowering}, a graph transformation method; and (5) HomoTTT~\cite{zhang2024fully}, an SSL-based fully TTT framework. For fair comparison, all methods are allocated the same tuning budget. Note that EERM triggered out-of-memory (OOM) errors on pubmed and arxiv using a 24GB NVIDIA 4090 GPU.
\begin{table}[t]
\caption{Comparison of in-distribution (ID) task performance before and after test-time training. The metric is the relative performance retention (\%).}
\label{Table:train_data}
\setlength{\tabcolsep}{0.3mm}{
\begin{tabular}{ccccc}
\toprule
 Dataset &  \multicolumn{1}{c}{Original Accuracy} & \multicolumn{3}{c}{Performance change (\%)} \\
\cmidrule(lr){2-2} \cmidrule(l){3-5}
(\%)  & Pre-trained Model   & \sys & Tent & HomoTTT \\ 
\midrule
cora     & 96.17            & 0.00      & -0.31  & -0.11   \\
pubmed   & 94.53            & 0.00      & -0.41  & -0.13   \\
citeseer & 99.34            & 0.00      & -21.13 & -25.32  \\
wikics   & 86.84            & 0.00      & -0.71  & -0.51 \\ 
\bottomrule
\end{tabular}}
\end{table}
\subsection{OOD Generalization Performance (RQ1)}
To evaluate \sys's effectiveness in OOD scenarios, we conduct a comprehensive comparison with five state-of-the-art OOD generalization methods. The accuracy results are summarized in Table~\ref{Table:main}, revealing three key findings: (1) \sys~consistently outperforms parameter finetuning methods across all datasets. This demonstrates that targeted adjustments to minimal node representations can effectively compensate for distribution shifts without the need for full parameter optimization. (2) While EERM outperforms others in most scenarios, its computational overhead limits practical deployment. This contrast highlights the advantage of post-hoc methods like \sys~that maintain plug-and-play simplicity without requiring expensive retraining of pre-trained models. (3) Robustness on Large-Scale Graphs: On the largest dataset, \sys~achieves improvement over the strong baseline TAR, proving its scalability and robustness in handling complex, real-world distribution shifts.

% \begin{figure}[t]
%   \centering
%   \begin{subfigure}[b]{0.4\textwidth}  % 稍微调大宽度，利用空间
%     \centering
%     \includegraphics[width=\linewidth]{Figures/ssl_ablation_study.pdf}
%     \caption{Ablation Study on Self-Supervised Tasks.}
%     \label{fig:ablation_SSL}
%   \end{subfigure}
%   \hfill
%   \begin{subfigure}[b]{0.4\textwidth}
%     \centering
%     \includegraphics[width=\linewidth]{Figures/node_intervention_ablation.pdf}
%     \caption{Ablation Study on Node Selection Strategies.}
%     \label{fig:ablation_intervention}
%   \end{subfigure}
%   \caption{Ablation studies on key components.
%   (a) Comparison of different self-supervised objectives for guiding intervention optimization. 
%   (b) Impact of node selection strategies on pubmed dataset. The bars represent accuracy when applying interventions to: no nodes (baseline), all nodes, test nodes, and our uncertainty-guided subset (top 10\% entropy).}
%   \label{fig:ablation}
% \end{figure}

\begin{figure}[t]
  \centering
    \includegraphics[width=\linewidth]{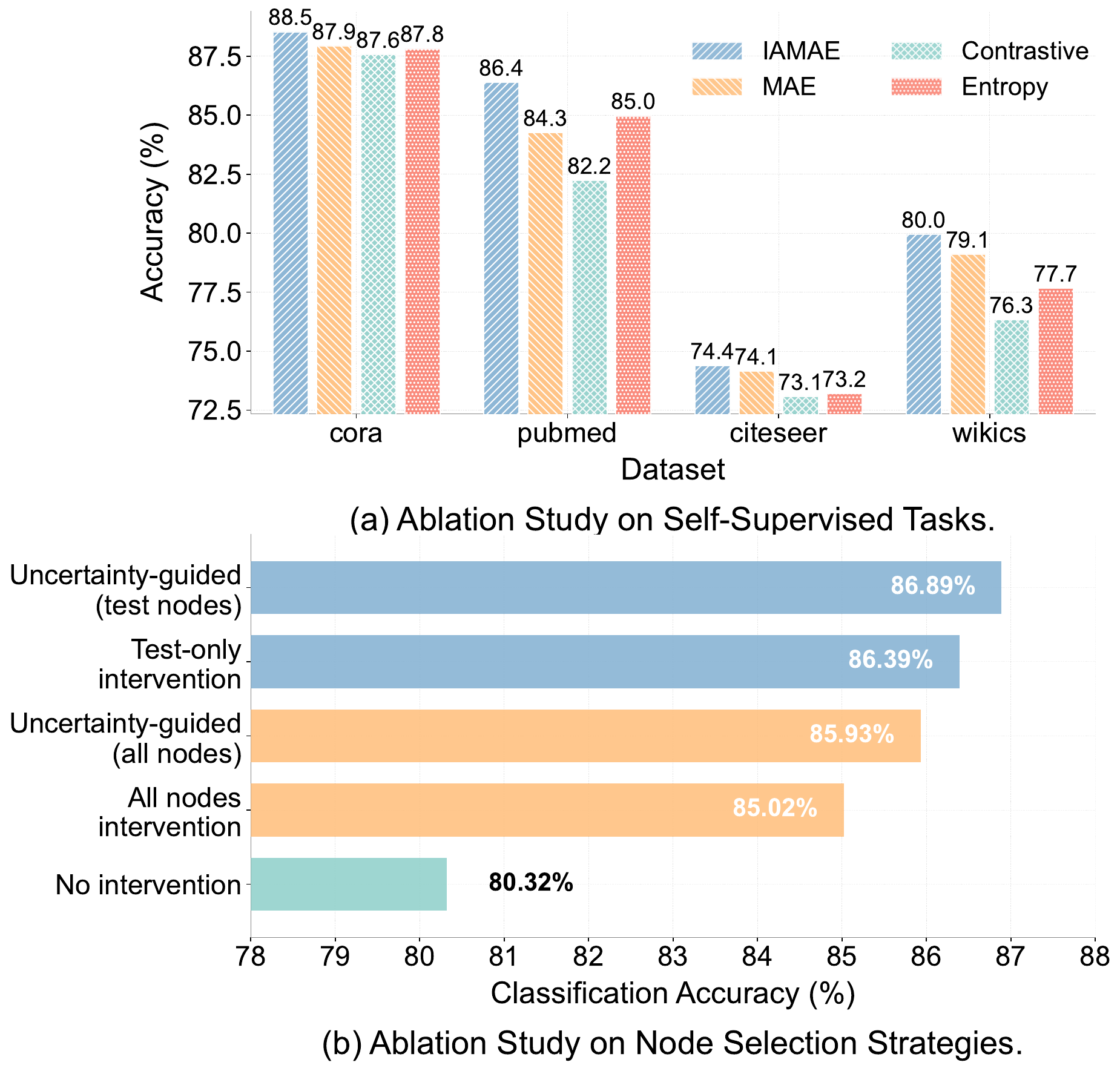}
  \caption{Ablation studies on key components.
  (a) Comparison of different self-supervised objectives for guiding intervention optimization. 
  (b) Impact of node selection strategies on pubmed dataset. The bars represent accuracy when applying interventions to: no nodes (baseline), all nodes, test nodes, and our uncertainty-guided subset (top 10\% entropy).}
  \label{fig:ablation}
\end{figure}
\subsection{Catastrophic Forgetting Analysis (RQ2)}
We quantitatively evaluate catastrophic forgetting by measuring the performance change on in-distribution (ID) data after test-time training. As shown in Table~\ref{Table:train_data}, our framework preserves 100\% of its original accuracy. This is structurally guaranteed by our design, which freezes the pre-trained parameters and confines adaptation to the representation space. In sharp contrast, parameter finetuning baselines suffer significant degradation due to catastrophic forgetting, rendering the adapted models unusable for the original task. 
% This validates that \sys effectively resolves the stability-plasticity dilemma inherent in conventional TTT.

\subsection{Ablation Study (RQ3)}
% \begin{figure}[t]
%     \centering
%     \includegraphics[width=1.0\linewidth]{Figures/ablation_study.png}
%     \caption{Performance comparison of different self-supervised tasks used to guide the optimization of the learnable interventions.}
%     \label{fig:SSL}
% \end{figure}

We validate two key components of TTReFT: the Intervention-Aware Masked Autoencoder and the Uncertainty-Guided Node Selection.
Firstly, we compare IAMAE against three baselines: standard MAE, Contrastive learning, and Entropy minimization. As shown in Figure~\ref{fig:ablation}(a), IAMAE consistently outperforms all variants across four datasets. This superiority stems from its dynamic masking strategy, which forces the model to reconstruct structurally significant features in high-intervention regions rather than learning trivial patterns.
Then, we evaluate different selection strategies on pubmed in Figure~\ref{fig:ablation}(b). Our Uncertainty-Guided approach achieves the highest accuracy, significantly surpassing Random Selection and No Intervention. Notably, intervening on Test-only nodes yields better results than All nodes, suggesting that adapting training nodes introduces unnecessary noise (over-adaptation). These results confirm that prioritizing uncertain test nodes effectively targets samples most affected by distribution shifts.
\begin{figure}[t]
    \centering
    \includegraphics[width=\linewidth]{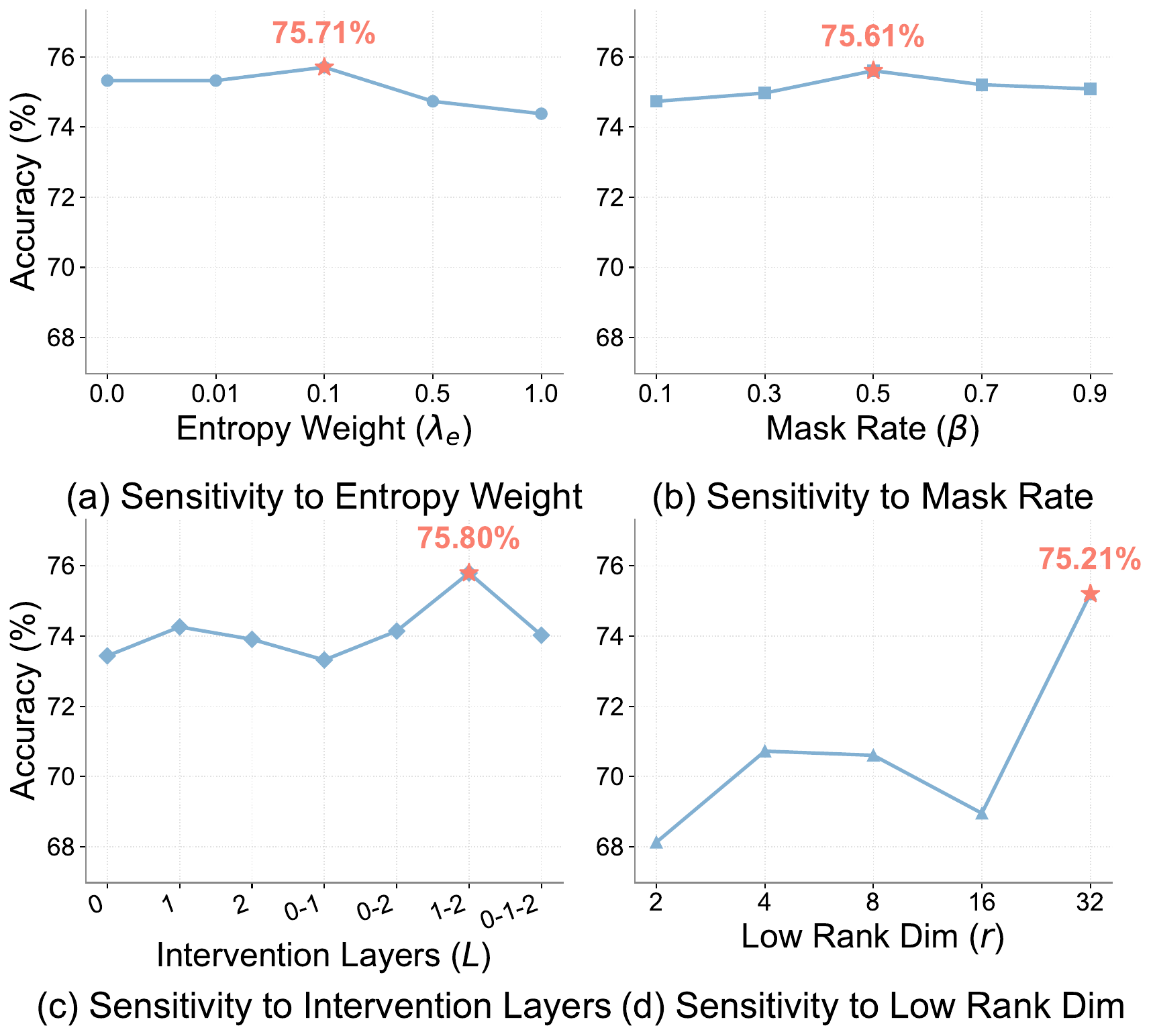}
    \caption{Sensitivity analysis of hyperparameters on the citeseer dataset. The \textcolor{red}{$\star$} marks the optimal value for each parameter.}
    \label{fig:sensitivity}
\end{figure}

\subsection{Sensitivity Analysis (RQ4)}
We examine the impact of four hyperparameters on citeseer, as shown in Figure~\ref{fig:sensitivity}.
(1)~\textbf{Robustness of entropy weight ($\lambda_e$) and mask rate ($\beta$)}. The model exhibits high robustness to both the $\lambda_e$ and $\beta$. Performance remains stable for $\lambda_e \in [0.0, 0.1]$ and $\beta \in [0.1, 0.9]$, peaking at $\lambda_e=0.1$ and $\beta=0.5$. Excessive entropy weight degrades performance by inducing over-confidence, while extreme mask rates disrupt the balance between self-supervision signals and structural context preservation.
(2)~\textbf{Intervention Layers.} Our results confirm that intervening on deeper layers significantly outperforms shallow ones, with multi-layer intervention yielding the best accuracy. Generally, deeper pre-trained backbones require interventions across more layers to effectively correct shifted representations as shown in Table~\ref{Table: hyperparameter} in Appendix.
(3)~\textbf{Low-Rank Dimension.} The optimal rank $r$ varies significantly across datasets, reflecting the complexity of the required adaptation. On citeseer, performance improves as $r$ increases, indicating that a higher-capacity intervention is needed to correct significant distribution shifts. Conversely, for datasets like cora and pubmed (detailed in Appendix), a minimal rank ($r=4$) is sufficient. This suggests that $r$ should be tuned to balance the expressiveness required for adaptation against the risk of overfitting, rather than simply maximizing the rank.
\begin{figure}[t]
    \centering
    \includegraphics[width=\linewidth]{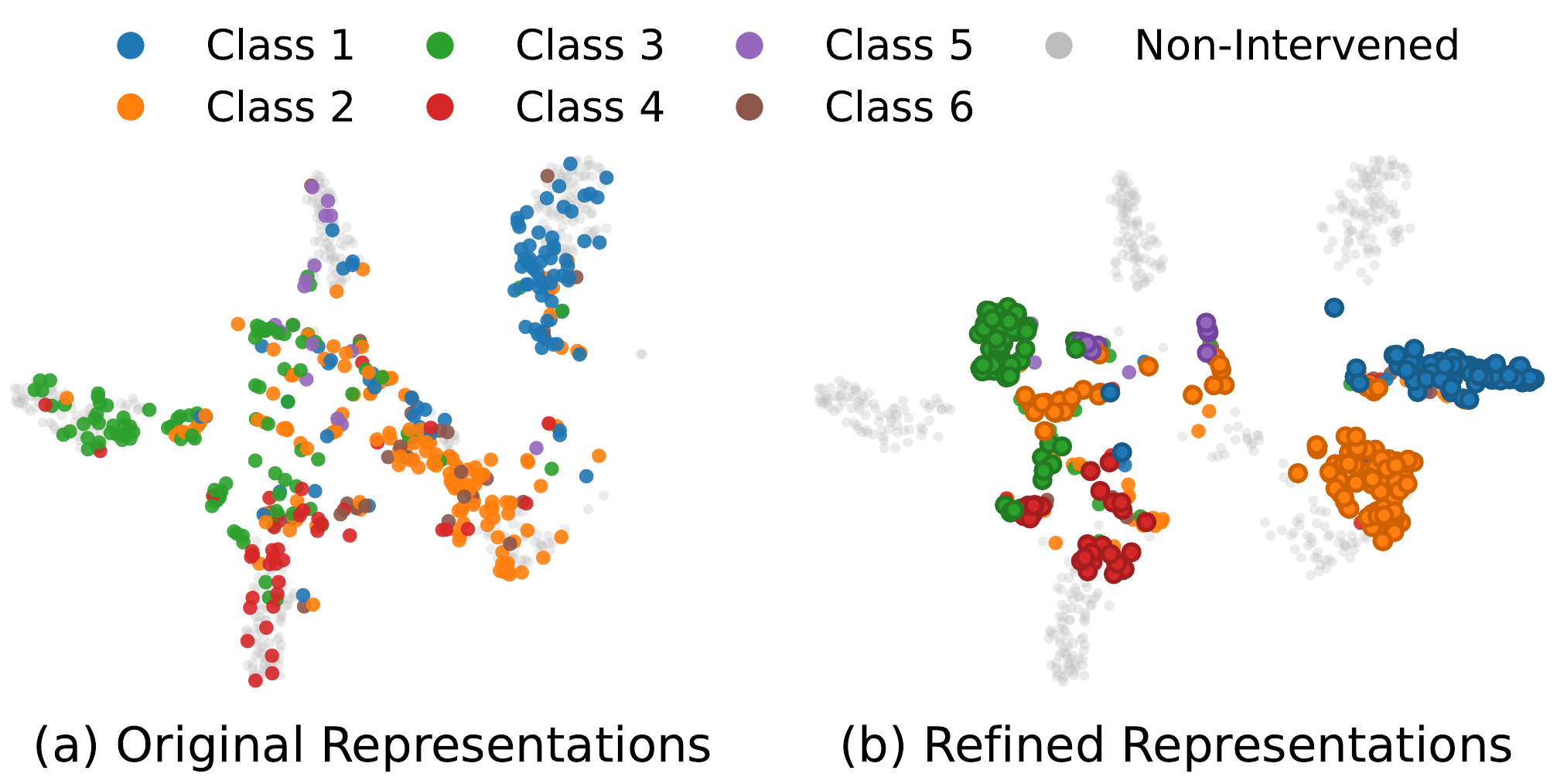}
    \caption{Visualization of representation refinement using t-SNE.}
    \label{fig:visualization}
\end{figure}
\subsection{Visualization}
To verify the efficacy of \sys, we visualize citeseer node representations before and after adaptation in Figure~\ref{fig:visualization}. Initially, nodes subject to distribution shift (colored) exhibit considerable overlap across class boundaries, leading to misclassification. After adaptation, these nodes significantly shift towards their ground-truth clusters, forming compact and distinct boundaries. This confirms that our representation finetuning effectively corrects shifted features by realigning them with the in-distribution manifold, directly improving classification accuracy.

\section{Related Work}

The related works of \sys~are discussed as follows:
\subsection{Graph OOD Generalization}
Graph OOD generalization aims to learn GNNs robust to distribution shifts. 
Invariant Learning approaches seek to capture stable substructures across environments. EERM~\cite{wu2022handling} and OOD-GNN~\cite{li2022ood} minimize risk variance or remove spurious correlations to achieve invariance. SR-GNN~\cite{zhu2022shift} employs adversarial clustering to enforce class separation under shifts.
Augmentation methods enhance robustness by diversifying training data. GSAT~\cite{miao2022interpretable} extracts robust subgraphs by filtering environmental bias, while TAR~\cite{zhengtopology} uses topology-aware reweighting to handle structural shifts. 
While effective, these methods heavily rely on diverse training domains or strong assumptions to simulate OOD patterns during training. In contrast, \sys~adopts a test-time training paradigm, enabling on-the-fly adaptation to unknown shifts without revisiting source data.

\subsection{Test-Time Training on Graphs}
Test-time training~\cite{sun2020test} has emerged as a promising paradigm for addressing out-of-distribution (OOD) challenges. 
GT3~\cite{wang2022test} pioneered TTT for graph neural networks via contrastive learning. 
Subsequent approaches introduce node-level adaptations: HomoTTT leverages homophily priors~\cite{zhang2024fully}, while LLMTTT incorporates language model annotations~\cite{zhang2024test}.
However, these methods fundamentally suffer from catastrophic forgetting due to gradient conflicts~\cite{liu2021ttt++}. Our work circumvents these issues through a novel representation-space adaptation paradigm.

\subsection{Representation Finetuning}
Representation finetuning shifts the adaptation paradigm from adjusting model parameters to directly intervening on hidden representations. Pioneered by ReFT~\cite{wu2024reft}, which shows that such interventions can match or exceed full finetuning performance, this approach is further advanced by RED~\cite{wu2024advancing} via targeted network edits for high parameter efficiency. By completely decoupling adaptation from parameter updates, these methods effectively preserve pre-trained knowledge. However, existing techniques are graph-agnostic. Our work is the first to systematically introduce the representation fine-tuning paradigm into the graph domain

% \subsection{Graph Autoencoders (GAEs)}
% Generative self-supervised learning on graphs has evolved through three phases: structural reconstruction (e.g., AGE~\cite{cui2020adaptive}, VGAE~\cite{kipf2016variational}), joint feature-structure reconstruction (e.g., MGAE~\cite{wang2017mgae}, GALA~\cite{park2019symmetric}, GATE~\cite{salehi2019graph}), and more recently, feature-focused reconstruction as in GraphMAE~\cite{hou2022graphmae}. Although masked reconstruction benefits node-level tasks, its random masking is suboptimal for test-time adaptation. We address this via uncertainty-guided masking that focuses on distributionally shifted regions, avoiding the augmentation reliance of contrastive methods like GraphCL~\cite{you2020graph} and other related frameworks~\cite{hu2019strategies}.

\section{Conclusion}

% This paper introduces \sys, a novel test-time representation finetuning framework designed to overcome the critical challenge of catastrophic forgetting in test-time training. \sys~achieves this by employing representation interventions as an alternative to direct parameter updates. This approach achieves robust out-of-distribution generalization through targeted adaptations while completely preserving original task knowledge.
% \sys~establishes a principled methodology for TTT through its uncertainty-guided intervention strategy and intervention-aware masked reconstruction loss. Functioning as an efficient, modular component, \sys~operates without requiring architectural modifications to the base model. Comprehensive experiments across diverse benchmarks and theoretical analysis demonstrate the effectiveness of our representation-centric paradigm for test-time adaptation. Beyond advancing TTT techniques specifically for graph data, our findings establish a foundation for future research into handling dynamic graph scenarios and scaling to deeper architectures via principled representation interventions.
This paper presents \sys, a novel test-time representation finetuning framework designed to circumvent the persistent challenge of catastrophic forgetting in test-time training (TTT). By shifting the adaptation paradigm from direct parameter updates to targeted representation interventions, \sys~achieves robust out-of-distribution generalization while strictly preserving pre-trained knowledge.
We establish a principled methodology for graph TTT, characterized by an uncertainty-guided node selection strategy and an intervention-aware masked reconstruction objective. As a lightweight, modular component, \sys~seamlessly integrates with existing architectures without necessitating structural modifications. 
Both extensive empirical evaluations across diverse benchmarks and theoretical analysis confirm the efficacy of this representation-centric approach. Beyond advancing TTT techniques for graph data, our work paves the way for future research into dynamic graph adaptation and scalable deep learning through the lens of principled representation refinement.

\begin{acks}
To Robert, for the bagels and explaining CMYK and color spaces.
\end{acks}

\bibliographystyle{ACM-Reference-Format}
\bibliography{sample-base}

\newpage
\appendix
\newpage
\section{Datasets}
\label{Appendix:Datasets}

This paper utilizes popular datasets commonly used for node classification tasks, namely cora~\cite{mccallum2000automating}, pubmed~\cite{sen2008collective}, citeseer~\cite{10.1145/276675.276685}, arxiv~\cite{hu2021open}, and wikics~\cite{mernyei2022wikics}. Afterwards, we apply the GOOD~\cite{gui2022good} split method to partition the aforementioned datasets. 
GOOD make distinctions between concept shifts and covariate shifts. We select degree and word (time in arxiv) as the criteria for domain division. According to different split methods, we generated four out-of-distribution (OOD) datasets for each dataset.
\begin{table}[t]
\caption{The description of datasets. A summary of the key statistics for the node classification datasets used in our experiments, including the number of classes, nodes, edges, and test nodes. The datasets are partitioned using the GOOD benchmark to explicitly create out-of-distribution splits based on criteria like node degree and feature (word).}
\label{Table: Datasets}
\setlength{\tabcolsep}{2.8mm}{
\begin{tabular}{ccccc}
\toprule
Dataset    & Class & Nodes  & Edges & Test nodes \\  \midrule
cora       & 7     & 2708   &  5429    & 837           \\
pubmed     & 3     & 19717  & 44338 & 6001        \\
citeseer   & 6     & 3186   & 4732 & 847         \\
wikics     & 10    & 11701  & 216123 & 3308         \\
arxiv & 40    & 169343 & 1166243 & 51480       \\ \bottomrule
\end{tabular}}
\end{table}
\section{Further Experimental Studies}

\subsection{Ablation on the Decoder Architecture}
\begin{table}[h]
\caption{The ablation of the decoder architecture. Node classification accuracy (\%) for TTReFT using different decoder architectures in the IAMAE module. The GCN-based decoder, which leverages graph structural information for feature reconstruction, consistently yields the best performance, validating its design for graph-structured data.}
\label{Table: ablation_decoder}
\begin{tabular}{cccc}
\toprule
Dataset  & GCN\_based & MLP\_based & Linear\_based \\ \midrule
cora     & \textbf{88.77}      & 87.81      & 88.05         \\
pubmed   & \textbf{86.89}      & 84.15      & 85.78         \\
citeseer & \textbf{75.80}      & 74.69      & 74.80          \\
wikics   & \textbf{79.96}      & 76.75      & 76.15        \\
\bottomrule
\end{tabular}
\end{table}
The decoder $f_D$ maps the latent code $H$ back to the input $X$, and its design would depend on the semantic level~\cite{he2022masked} of target $X$. In graphs, the decoder reconstructs relatively less informative multi-dimensional node features. In our proposed \sys, IAMAE uses GCN~\cite{kipf2017semisupervised} as the decoder. To have a fair comparison and investigate the influence of the backbone of decoder, we compare the results of different datasets using GCN, MLP and Linear as the decoder. The results are shown in Table~\ref{Table: ablation_decoder}. It demonstrates that with the GNN decoder, a masked node is forced to reconstruct its
input feature from the neighboring unmasked latent representations.
\begin{table}[t]
\caption{The ablation on the representation intervention instances. Performance comparison of different intervention mechanisms: our proposed LoReFT, a simplified variant without orthogonality constraints (DiReFT), and a minimal baseline using UV-decomposition (UV). The results demonstrate the importance of the orthogonal projection and residual design in LoReFT for effective and stable representation adaptation.}
\label{Table: ablation_intervention}
\centering
\begin{tabular}{cccc}
\toprule
Dataset  & LoReFT & DiReFT & UV    \\ \midrule
cora     & \textbf{88.77}  & 87.14  & 88.03 \\
pubmed   & \textbf{86.89}  & 83.67  & 83.34 \\
citeseer & \textbf{75.80}  & 74.48  & 74.31 \\
wikics   & \textbf{79.96}  & 76.94  & 77.32 \\
\bottomrule
\end{tabular}
\end{table}
\subsection{Ablation on the Representation Intervention Instance}
To investigate the impact of intervention mechanisms on model representation finetuning, we conduct ablation studies centered on the low-rank linear subspace finetuning method LoReFT. Three comparative variants are constructed:  

\noindent \textbf{Baseline Method (LoReFT):}  
   Following the original formulation, LoReFT intervenes hidden representations within an $r$-dimensional subspace spanned by the row vectors of matrix $R$ through the projection source $\mathbf{Rs = Wh + b}$. Parameters $\{\mathbf{R, W, b}\}$ are learnable while the pre-trained model remains frozen. Critically, $\mathbf{R} \in \mathbb{R}^{r \times d}$ is enforced as row-orthogonal to ensure subspace regularity and mitigate representation collapse.  

\noindent \textbf{Simplified Variant (DiReFT):  }
   We remove LoReFT’s orthogonality constraint and residual mechanism, reducing it to $\Phi_{\text{DiReFT}}(\mathbf{h}) = \mathbf{h} + \mathbf{W}_2^\top (\mathbf{W}_1 \mathbf{h} + \mathbf{b})$. Here, $\mathbf{W}_1, \mathbf{W}_2 \in \mathbb{R}^{r \times d}$ are unconstrained low-rank projection matrices. This variant reduces computational overhead and aligns with direct hidden-state interventions like LoRA~\cite{hu2022lora}.  

\noindent \textbf{Control Method (UV-Decomposition):  }
   As a minimal-intervention baseline, we introduce elementary low-rank matrix factorization: $\Phi_{\text{UV}}(\mathbf{h}) = \mathbf{h} + \mathbf{U}\mathbf{V}^\top \mathbf{h}$, where $\mathbf{U}, \mathbf{V} \in \mathbb{R}^{d \times r}$. This approach applies no subspace constraints or nonlinear projections.

This ablation isolates the effects of (i) orthogonality constraints, (ii) residual design, and (iii) projection complexity on low-rank adaptation efficacy. Table~\ref{Table: ablation_intervention} shows the effectiveness of our intervention mechanisms.
\begin{table*}[t]
\caption{Empirical efficiency comparison of test-time adaptation methods. Metrics were measured on the PubMed dataset. \sys achieves a superior trade-off by optimizing a minimal set of low-rank parameters.}
\label{Table:efficiency_comparison}
\centering
\begin{tabular}{ccccc}
\toprule
\text{Method} & \text{Adaptation Type} & \text{Avg. Time (s)} $\downarrow$ & \text{Peak Memory (GB)} $\downarrow$ & \text{Tunable Parameters} \\
\midrule
EERM & Environment Inference & OOM & $>$24 & $O(\text{\#environments})$ \\
Tent & Parameter Modulation & 28.4 & 6.5 & $\mathbf{\gamma, \beta} \in \mathbb{R}^d$ \\
HomoTTT & Full Parameter Finetuning & 35.1 & 7.2 & $\sim${100\% of $\theta$} \\
\textbf{TTReFT (Ours)} & \textbf{Representation Intervention} & \textbf{5.8} & \textbf{2.1} & $\mathbf{R}, \mathbf{W}, \mathbf{b}$\\
\bottomrule
\end{tabular}
\end{table*}
\subsection{Efficiency Comparison}
\label{Appen:Efficiency comparison}
Beyond its strong out-of-distribution generalization performance, the proposed \sys framework achieves a significant improvement in test-time efficiency compared to traditional parameter finetuning (PaFT) methods. As shown in Table~\ref{Table:efficiency_comparison}, efficiency metrics are measured on the PubMed dataset during the test-time tarining phase. A higher Avg. Time indicates faster adaptation. The number of tunable parameters for \sys is determined by the low-rank dimension $r$  and representation dimension $d$, which is independent of the model size. \sys requires substantially less computation time and memory during the test-time adaptation phase, thanks to its parameter-efficient representation intervention strategy.

While PaFT methods must compute gradients and update a large portion of the model's parameters, \sys only optimizes a small set of low-rank intervention matrices while keeping all pre-trained parameters frozen. This approach reduces the number of tunable parameters by orders of magnitude—from $O(Nd^2)$ to $O(|P|rd)$—resulting in faster convergence and lower memory footprint. The efficiency gains are particularly pronounced on larger graphs, where \sys maintains stable memory usage while PaFT baselines often encounter out-of-memory (OOM) errors. These advantages make \sys highly practical for real-world deployment where test-time computational resources are often limited.
\section{Implementation Details}

\begin{table*}[t]
\caption{Hyperparameter configurations of the proposed \sys framework across different datasets. The table presents the optimal values found for each key parameter during validation, along with their respective tuning ranges.}
\label{Table: hyperparameter}
\setlength{\tabcolsep}{1.1mm}{
\begin{tabular}{@{}ccccccccc@{}}
\toprule
\multirow{2}{*}{Hyperarameter} & \multirow{2}{*}{Symbol} & \multicolumn{5}{c}{Optimal Value by Dataset} & \multirow{2}{*}{Search Space} & \multirow{2}{*}{Description} \\
\cmidrule(l){3-7}
 & & cora & pubmed & citeseer & wikics & arxiv  \\
\midrule
Low-rank dimension & $r$ & 8 & 32 & 32 & 4 & 2 & $\{2,4,8,16,32\}$ & Rank of intervention subspace \\
Intervention layers & $L$ & 1 & $[1,2]$ & $[2,3]$ & 2 & 1 & $\{1, 2, \dots, K\}$ & Layers to intervene \\
Learning rate & $\eta$ & 0.01 & 0.01 & 0.0003 & 0.008 & 0.006 & $[10^{-4},10^{-2}]$ & Optimizer learning rate \\
Entropy loss weight & $\lambda_e$ & 0.1 & 0.1 & 0.1 & 0.1 & 0.1 &  fixed & Entropy loss coefficient \\
Gate temperature & $\alpha_\text{gate}$ & 10 & 10 & 10 & 10 & 10 & fixed & Intervention decision sharpness \\
Global mask rate & $\rho$ & 0.7 & 0.2 & 0.5 & 0.3 & 0.2 & [0, 0.8] & Global masking probability \\
Base mask rate & $\beta$ & 0.5 & 0.5 & 0.5 & 0.5 & 0.5 & fixed & Base masking probability \\
\bottomrule
\end{tabular}}
\end{table*}
\subsection{Environment}
All experiments(including main experiments and comparison experiments) are conducted on Windows servers equipped with an NVIDIA 4090 GPU. Models are implemented in PyTorch version 2.1.0 with CUDA version 12.1, scikit-learn version 1.3.2 and Python 3.8. 
\subsection{Model Configuration}
For node classification, we train the model using Adam Optimizer with $\beta_1 = 0.9, \beta_2 = 0.999, \varepsilon = 1×10-8$. More details about hyper-parameters and datasets are in Table~\ref{Table: hyperparameter}. The intervention layers $L$ can be either a single layer or any set of consecutive layers from the pre-trained GNN model, allowing flexible adaptation at different semantic levels.

\section{Further Theoretical Analysis}
\label{Appedix:Proof}
Let \(x, y \sim P\) denote the training distribution. Consider a node classification task with \(C\) classes using a \(K\)-layer Simplified Graph Convolution (SGC)~\cite{wu2019simplifying} model. The node representation is given by \(Z = A^K X W\), where \(A\) is the adjacency matrix, \(X\) is the feature matrix, and \(W \in \mathbb{R}^{n \times d}\) is the weight matrix. Assume the ground-truth label \(y\) is generated as \(y = \mathrm{softmax}(A x W)\) with \(K=1\).

At test time, we simulate distribution shift in the test samples using an orthogonal transformation \(Q\), and we observe shifted samples \((\tilde{x}, y) \sim P_t\) with \(\tilde{x} = Q x\).
We then apply a test-time intervention \(\Phi(\tilde{x}) = (1-\alpha)\tilde{x} + \alpha UV\tilde{x}\), where \(\alpha\) is a hyper-parameter, and predict \(\hat{y}_\alpha = \mathrm{softmax}(A \Phi(\tilde{x}) W)\). Define the prediction risk as \({Risk}(\alpha) = \mathbb{E}[\| \hat{y}_\alpha - y \|_1]\), where the expectation is taken over \(P_t\).

To facilitate the subsequent proof, we state the following assumptions:
\begin{enumerate}
    \item The intervention operation is effective, i.e., \(\mathbb{E}[\| (UV Q - I) x \|_2^2] < \mathbb{E}[\| (Q - I) x \|_2^2]\). This indicates that the intervention effect after processing with \(UV\) is superior to that without \(UV\) processing.
    \item The training feature \(x\) is bounded, satisfying \(\|x\|_2 \leq M\) almost surely. Its expectation is zero (\(\mathbb{E}[x] = 0\)) and its covariance matrix is \(\Sigma_x\) (\(\mathrm{Cov}(x) = \Sigma_x\)).
    \item The softmax function satisfies Lipschitz continuity, i.e., there exists a constant \(L > 0\) such that \(\| \mathrm{softmax}(z) - \mathrm{softmax}(z') \|_1 \leq L \|z - z'\|_2\). This implies that the \(\ell_1\)-norm difference of the softmax output is controlled by the \(\ell_2\)-norm difference of the input with a coefficient \(L\).
    \item The test distribution \(P_t\) is the distribution of \((QX, Y)\) where \((X,Y) \sim P\).
\end{enumerate}

Under these assumptions, there exists \(\alpha > 0\) such that \({Risk}(\alpha) < {Risk}(0)\).

\noindent  \textbf{Proof.}
We establish the theorem by demonstrating that the test-time intervention reduces the expected deviation between the true logits and the predicted logits, which subsequently lowers the prediction risk due to the Lipschitz continuity of the softmax function. The proof proceeds through five interconnected stages.

1. Logit Formulation and Residual Analysis

Let \(x\) denote the feature vector of a test node. The ground-truth logits are given by:
\begin{equation*}
z_{\mathrm{true}} = A x W.
\tag{1}
\end{equation*}
Under test-time distribution shift, we observe \(\tilde{x} = Q x\), where \(Q\) is an orthogonal matrix. The test-time intervention transforms the shifted features as:
\begin{equation}
\Phi(\tilde{x}) = (1 - \alpha) \tilde{x} + \alpha UV \tilde{x} = \left[ (1 - \alpha) Q + \alpha UV Q \right] x.
\tag{2}
\end{equation}
The resulting logits are:
\begin{equation}
z_\alpha = A \Phi(\tilde{x}) W = A \left[ (1 - \alpha) Q + \alpha UV Q \right] x W.
\tag{3}
\end{equation}
Define the residual corruption matrix \(C = Q - I\) and the residual repair matrix \(D = U V Q - I\). The deviation between the intervened logits and the true logits is:
\begin{align}
z_\alpha - z_{\mathrm{true}} &= A \left[ (1 - \alpha) Q + \alpha UV Q - I \right] x W \nonumber \\
&= A \left[ (1 - \alpha) (Q - I) + \alpha (UV Q - I) \right] x W \nonumber \\
&= A \left[ (1 - \alpha) C + \alpha D \right] x W.
\tag{4}
\end{align}

2. Expected Squared Distance Computation

To quantify the deviation, we analyze the expected squared Euclidean distance between the logits:
\begin{equation}
d(\alpha) = \mathbb{E} \left[ \| z_\alpha - z_{\mathrm{true}} \|_2^2 \right].
\tag{5}
\end{equation}
Express the squared distance as:
\begin{align}
\| z_\alpha - z_{\mathrm{true}} \|_2^2 &= \left( A S_\alpha x W \right)^\top \left( A S_\alpha x W \right),
\tag{6}
\end{align}
where \(S_\alpha = (1 - \alpha) C + \alpha D\). Utilizing properties of the Kronecker product:
\begin{align}
(A S_\alpha x W) &= (W^\top \otimes A) (S_\alpha x),
\tag{7}
\end{align}
yields:
\begin{align}
\| z_\alpha - z_{\mathrm{true}} \|_2^2 &= (S_\alpha x)^\top (W W^\top \otimes A^\top A) (S_\alpha x).
\tag{8}
\end{align}
Denoting \(Q = W W^\top \otimes A^\top A\), and noting that \((S_\alpha x) = (I \otimes S_\alpha) x\), we obtain:
\begin{align}
d(\alpha) &= \mathbb{E} \left[ x^\top (I \otimes S_\alpha^\top) Q (I \otimes S_\alpha) x \right].
\tag{9}
\end{align}
Given that \(\mathbb{E}[x] = 0\) and \(\mathbb{E}[x x^\top] = \Sigma_x \otimes I\), we apply the trace expectation formula:
\begin{align}
d(\alpha) &= \mathrm{tr} \left( (I \otimes S_\alpha^\top) Q (I \otimes S_\alpha) (\Sigma_x \otimes I) \right).
\tag{10}
\end{align}
Using trace identities for Kronecker products:
\begin{align}
d(\alpha) &= \mathrm{tr}(S_\alpha^\top A^\top A S_\alpha \Sigma_x) \cdot \mathrm{tr}(W W^\top).
\tag{11}
\end{align}
Let \(\gamma = \mathrm{tr}(W W^\top) > 0\) (since \(W \neq \mathbf{0}\)), resulting in:
\begin{equation}
d(\alpha) = \gamma \cdot \mathrm{tr}(S_\alpha^\top A^\top A S_\alpha \Sigma_x).
\tag{12}
\end{equation}

3. Quadratic Parameterization and Coefficient Analysis

Substituting \(S_\alpha = (1 - \alpha) C + \alpha D\) into (12) and expanding:
\begin{align}
d(\alpha) &= \gamma \left[ (1 - \alpha)^2 \mathrm{tr}(C^\top A^\top A C \Sigma_x) \right. \nonumber \\
&\quad + \alpha(1 - \alpha) \left( \mathrm{tr}(C^\top A^\top A D \Sigma_x) + \mathrm{tr}(D^\top A^\top A C \Sigma_x) \right) \nonumber \\
&\quad + \left. \alpha^2 \mathrm{tr}(D^\top A^\top A D \Sigma_x) \right].
\tag{13}
\end{align}
Define the coefficients:
\begin{equation}
\begin{split}
E &= \mathrm{tr}(C^\top A^\top A C \Sigma_x), \\
F &= \mathrm{tr}(C^\top A^\top A D \Sigma_x) + \mathrm{tr}(D^\top A^\top A C \Sigma_x), \\
G &= \mathrm{tr}(D^\top A^\top A D \Sigma_x),
\end{split}
\tag{14}
\end{equation}
simplifying (13) to:
\begin{equation}
d(\alpha) = \gamma \left[ (1 - \alpha)^2 E + \alpha(1 - \alpha) F + \alpha^2 G \right].
\tag{15}
\end{equation}
Algebraic expansion reveals the quadratic form:
\begin{align}
d(\alpha) &= \gamma \left[ (1 - 2\alpha + \alpha^2) E + (\alpha - \alpha^2) F + \alpha^2 G \right] \nonumber \\
&= \gamma \left[ E + \alpha(-2E + F) + \alpha^2(E - F + G) \right] \nonumber \\
&= \gamma \left( a \alpha^2 + b \alpha + c \right),
\tag{16}
\end{align}
where \(a = E - F + G\), \(b = F - 2E\), and \(c = E\).

4. Efficacy of the Test-Time Intervention

The baseline (\(\alpha = 0\)) and full-intervention (\(\alpha = 1\)) distances are:
\begin{align}
d(0) &= \gamma E = \gamma \cdot \mathbb{E} \left[ \| A (Q - I) x \|_2^2 \right],
\tag{17} \\
d(1) &= \gamma (a + b + c) = \gamma G = \gamma \cdot \mathbb{E} \left[ \| A (UV Q - I) x \|_2^2 \right].
\tag{18}
\end{align}
By the intervention effectiveness assumption (Assumption 1):
\begin{equation}
\mathbb{E} \left[ \| (UV Q - I) x \|_2^2 \right] < \mathbb{E} \left[ \| (Q - I) x \|_2^2 \right],
\tag{19}
\end{equation}
and since \(A^\top A \succeq 0\) (positive semi-definite) and \(\gamma > 0\), we have:
\begin{equation}
d(1) < d(0).
\tag{20}
\end{equation}
As \(d(\alpha)\) is continuous in \(\alpha\), and \(d(1) < d(0)\), there exists \(\alpha^* \in (0, 1]\) such that:
\begin{equation}
d(\alpha^*) < d(0).
\tag{21}
\end{equation}

5. Risk Minimization via Lipschitz Continuity

The prediction risk is bounded using the Lipschitz continuity of softmax:
\begin{align}
{Risk}(\alpha) &= \mathbb{E} \left[ \| \mathrm{softmax}(z_\alpha) - \mathrm{softmax}(z_{\mathrm{true}}) \|_1 \right] \nonumber \\
&\leq L \cdot \mathbb{E} \left[ \| z_\alpha - z_{\mathrm{true}} \|_2 \right]. \quad (\text{by Assumption 3})
\tag{22}
\end{align}
By Jensen's inequality for the concave square root function:
\begin{align}
\mathbb{E} \left[ \| z_\alpha - z_{\mathrm{true}} \|_2 \right] \leq \sqrt{ \mathbb{E} \left[ \| z_\alpha - z_{\mathrm{true}} \|_2^2 \right] } = \sqrt{d(\alpha)},
\tag{23}
\end{align}
which implies:
\begin{equation}
{Risk}(\alpha) \leq L \sqrt{d(\alpha)}.
\tag{24}
\end{equation}
For \(\alpha = \alpha^*\), we have:
\begin{equation}
{Risk}(\alpha^*) \leq L \sqrt{d(\alpha^*)} < L \sqrt{d(0)}.
\tag{25}
\end{equation}
To establish the strict inequality for the actual risk, observe that the mapping \(z \mapsto \| \mathrm{softmax}(z) - y \|_1\) is continuous and (locally) convex in a neighborhood of \(z_{\mathrm{true}}\) (or at least, the risk is minimized at \(z_{\mathrm{true}}\)). Since \(d(\alpha^*) < d(0)\) implies:
\begin{equation}
\mathbb{E} \left[ \| z_{\alpha^*} - z_{\mathrm{true}} \|_2^2 \right] < \mathbb{E} \left[ \| z_0 - z_{\mathrm{true}} \|_2^2 \right],
\tag{26}
\end{equation}
and the risk is minimized at \(z_{\mathrm{true}}\), we conclude:
\begin{equation}
{Risk}(\alpha^*) < {Risk}(0),
\tag{27}
\end{equation}
completing the proof.

\noindent \textbf{Remark.} The optimal intervention parameter \(\alpha^*\) minimizes the quadratic function \(d(\alpha)\). If \(a > 0\), it can be computed as \(\alpha^* = -\frac{b}{2a}\). In practice, the intervention hyper-parameter \(\alpha\) may be tuned on a validation set to maximize repair efficacy. The design of the low-rank intervention module (\(UV\)) is more complex; in our analysis, we simplified it to the matrix \(UV\), and this simplification does not affect the overall analytical framework. The orthogonal shift assumption simplifies the analysis but can be relaxed to invertible \(Q\) without loss of generality.

\end{document}